\documentclass{ecai} 
\usepackage{latexsym}
\usepackage{amssymb}
\usepackage{amsmath}
\usepackage{amsthm}
\usepackage{booktabs}
\usepackage{enumitem}
\usepackage{graphicx}
\usepackage{color}
\usepackage{multirow}

\newcommand{\BibTeX}{B\kern-.05em{\sc i\kern-.025em b}\kern-.08em\TeX}

\begin{document}

%%%%%%%%%%%%%%%%%%%%%%%%%%%%%%%%%%%%%%%%%%%%%%%%%%%%%%%%%%%%%%%%%%%%%%%%

\begin{frontmatter}

%%% Use this command to specify the title of your paper.

\title{Spatio-Temporal Graph Neural Networks for Infant Language Acquisition Prediction}

%%% Use this combinations of commands to specify all authors of your 
%%% paper. Use \fnms{} and \snm{} to indicate everyone's first names 
%%% and surname. This will help the publisher with indexing the 
%%% proceedings. Please use a reasonable approximation in case your 
%%% name does not neatly split into "first names" and "surname".
%%% Specifying your ORCID digital identifier is optional. 
%%% Use the \thanks{} command to indicate one or more corresponding 
%%% authors and their email address(es). If so desired, you can specify
%%% author contributions using the \footnote{} command.

\author[A]{\fnms{Andrew}~\snm{Roxburgh}\orcid{0000-0003-1333-5654}}
\author[A]{\fnms{Floriana}~\snm{Grasso}\orcid{0000-0001-8419-6554}}
\author[A]{\fnms{Terry R.}~\snm{Payne}\orcid{0000-0002-0106-8731}\thanks{Corresponding Author. Email: T.R.Payne@liverpool.ac.uk}} 

\address[A]{University of Liverpool\\
Liverpool, UK}

%%% Use this environment to include an abstract of your paper.

\begin{abstract}
Predicting the words that a child is going to learn next can be useful for boosting language acquisition, and such predictions have been shown to be possible with both neural network techniques (looking at changes in the vocabulary state over time) and graph model (looking at data pertaining to the relationships between words). However, these models do not fully capture the complexity of the language learning process of an infant when used in isolation. 
In this paper, we examine how a model of language acquisition for infants and young children can be constructed and adapted for use in a \emph{Spatio-Temporal Graph Convolutional Network (STGCN)}, taking into account the different types of linguistic relationships that occur during child language learning.  We introduce a novel approach for predicting child vocabulary acquisition, and evaluate the efficacy of such a model with respect to the different types of linguistic relationships that occur during language acquisition, resulting in insightful observations on model calibration and norm selection. 
An evaulation of this model found that the mean accuracy of models for predicting new words when using sensorimotor relationships (0.733) and semantic relationships (0.729) were found to be superior to that observed with a 2-layer Feedforward neural network. 
Furthermore, the high recall for some relationships suggested that some relationships (e.g. visual) were superior in identifying a larger proportion of relevant words that a child should subsequently learn than others (such as auditory).
\end{abstract}

\end{frontmatter}

%%%%%%%%%%%%%%%%%%%%%%%%%%%%%%%%%%%%%%%%%%%%%%%%%%%%%%%%%%%%%%%%%%%%%%%%
\section{Introduction}

\emph{Developmental Language Disorder} (DLD) is a condition whereby a child shows significant difficulties with language development for no clear reason. It is the most common disability in pre-schoolers, affecting somewhere between 5-7\% of all UK children \cite{lindsay_educational_2002-1,scerri_dcdc2_2011,tomblin_prevalence_1997}. The condition is exacerbated by other factors associated with learning difficulties, or environmental or familial factors.
For example, in the UK, between 40\% and 56\% of children in areas of social disadvantage begin nursery school with such a language delay \cite{law2011communication,locke_development_2002}. This has been proven to cause a knock-on effect within child education; for example, DLD has an incidence rate of 50-90\% on reading difficulties \cite{stothard_language_1998}, which can have a subsequent impact on educational outcomes (in fact reading levels have been shown to be a good predictor of educational outcomes \cite{hulme_developmental_2013}). Furthermore, children show great difficulties catching up with their peers without adequate support \cite{bleses_early_2016,conti-ramsden_follow-up_2001,feinstein_development_2006,van_dulm_does_2016}, and thus are more likely to perform worse academically, to show a lower employment rate, and to have poorer future mental health \cite{clegg_developmental_2005,snowling_educational_2001,young_young_2002}. 

Language delay is also an indicator of many other developmental issues; at least 3\% of all UK children have a language delay linked with hearing impairment, specific learning difficulties such as dyslexia, and general learning needs \cite{lindsay_educational_2002-1}.  Communication difficulties also feature in Autism Spectrum Disorder, which affects 1\% of the UK population \cite{baron-cohen_prevalence_2009}. Such challenges are not limited to neurologically `atypical' children, as the child's communication environment and family circumstances (e.g. poverty or limited language exposure) can have an effect on any early years language development, with a negative, cascading effect on school outcomes years later \cite{roulstone_investigating_2011,walker_prediction_1994}.

With such a public health issue of this scale, identifying affected children early is extremely important. One of the most commonly-used methods for identifying such children is through the use of standardised tools such as the \emph{MacArthur-Bates Communicative Development Inventory (CDI)} \cite{fenson_macarthur-bates_2007}, a paper instrument originally developed in US English but now adapted into many languages \cite{alcock_construction_2016}.  It consists of a series of questions and checklists designed to assess how children understand and produce words and gestures, as well as their acquisition of grammar.  It is typically accompanied by a Family Questionnaire, which gathers information about the family members and individuals who generally spend time with the child during a typical week. This data can be used to generate a \emph{spatial model of language acquisition} (where the spatial component corresponds to words with specific linguistic relationships), which can then be used to identify children who may suffer from language delay.

\begin{figure*}[t]
    \centering
        \includegraphics[width=0.7\textwidth]{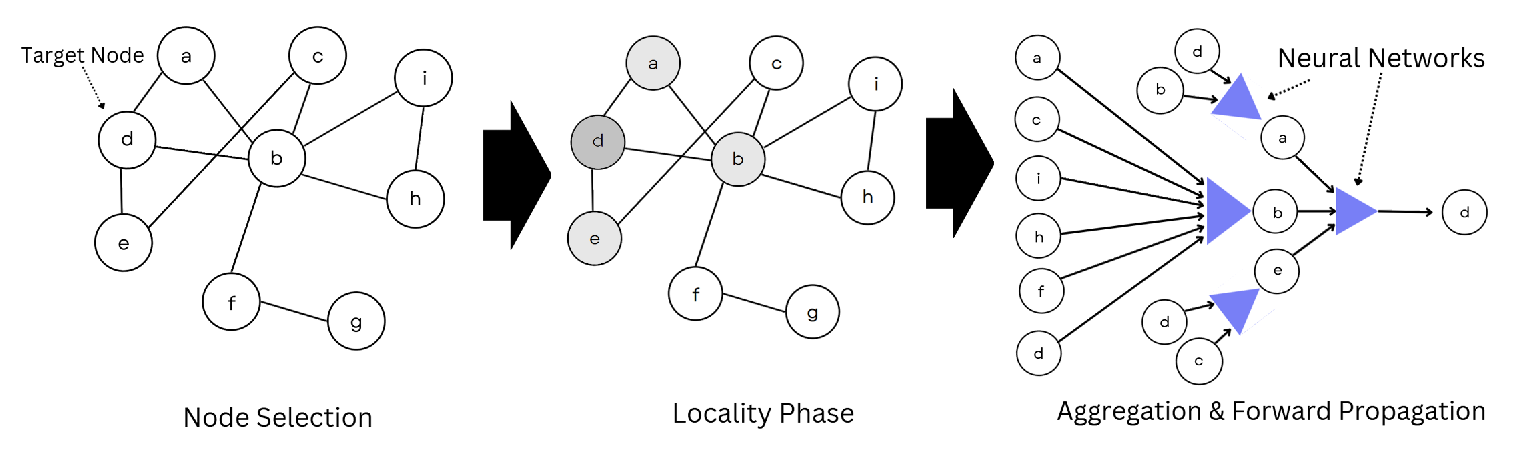}
        \caption{General operation of a \emph{Spatial Graph Neural Network} (GNN).}
        \label{fig:img1}
\end{figure*}

By automating the process of acquiring this data, through mobile applications such as \emph{Babytalk}\footnote{\url{http://www.lucid.ac.uk/}} \cite{roxburgh2017}, a model of a child's language acquisition can be built over time, and used to predict those words that the child should  learn in the future \cite{beckage_predicting_2015,beckage_quantifying_2020}. This knowledge can be used as a potential intervention for children with DLD, by suggesting words that a child's carer should emphasise when teaching language \cite{johnson_word_2001}; for example, using the \emph{Babytalk} application. 
This could also be used to boost the language acquisition of neurologically atypical children, and could have further research applications. 
One way to accomplish this is to refer to \emph{Age-of-Acquisition} norms \cite{stadthagen2006bristol} that can be used to encourage caregivers to emphasise words that typical children would acquire at a similar age. However, children (as individuals) do not tend to learn the same words at the same rate or age, and so a more tailored approach is required. The accuracy of the model used to predict forthcoming words is therefore an important factor.

\emph{Graph Neural Networks (GNN)} \cite{wu_surveyGNN_2021} extend the concept of conventional feedforward neural networks to capture the relationships and interactions between nodes within a graph, by  learning a vector representation of each node that not only depends on its own features but also on the features of its neighbours. These \emph{node embeddings} encode the structural and feature data of themselves, their neighbouring nodes, and ultimately extend to the overall graph structure. The resulting model can be used for edge prediction, node or graph classification, labelling, and feature prediction; or in the context of language acquisition, it could be used to determine whether the word that the node represents will be `known' by the child.

In this paper we examine how a model of language acquisition for infants and young children can be constructed and adapted for use in a \emph{Spatio-Temporal Graph Convolutional Network (STGCN)}, taking into account the different types of linguistic relationships that occur during child language learning.  We introduce a novel approach for attempting to predict child vocabulary acquisition, and evaluate the efficacy of such a model with respect to the different types of linguistic relationships that occur during language acquisition.
Section \ref{sec:existingWork} provides an overview of existing approaches that explore how vocabulary learning can be modelled, and provides a short overview of the STGCN based approach.  The way in which a child's vocabulary can be modelled as a graph is presented together with a discussion of how it is used by the STGCN in Section \ref{sec:model}. After describing the datasets collected for this study (Section~\ref{sec:data}), we evaluate the approach empirically (Section \ref{sec:results}) and conclude in Section \ref{sec:conclusions}.

%% %%%%%%%%%%%%%%%%%%%%%%%%%%%%%%%%%%%%%%%%%%%%%%%%%%%
\section{Background}
\label{sec:existingWork}

\subsection{Previous Approaches to Vocabulary Models}

Vocabulary acquisition can be thought of as the evolution of a network of interconnected nodes, where each node models a word and the edges represent a variety of different relationships between them (several of these relationships are discussed in Section \ref{sec:model}).  As more vocabulary is acquired,  the graph expands, and new connections are established. Much of the work in modelling child vocabulary growth has exploited the use of graph representations and artificial neural networks, due to the fact that the words a child learns are inherently connected with each other. Examples include the use of graphs when modelling vocabulary growth over time \cite{ke2008analysing}, and the use of neural networks for modelling the way that a brain acquires language \cite{sims2013exploring}. Thus, graph theory can be used as a means of analysing and simulating vocabularies and their evolution.  

Beckage, Mozer \& Colunga \cite{beckage_predicting_2015} showed that it is possible to predict words that would be learnt in the future using the words a child already knows.  By analysing CDI questionnaire data of 77 subjects over a 1-year period at 1-month intervals, and by using a network growth technique, they constructed three different models (\emph{Additive}, \emph{Maximal}, and \emph{Threshold}), each calculating the probability of a word being learnt within the next month in slightly different ways, but all three relying on the probability of a new word being learned as a consequence of the set of words already learnt, and specifically the other words learnt concurrently (i.e. during that same 1-month period). They also successfully developed several neural-network based predictive models using a variety of qualitatively different sources of information as inputs, and observed that the accuracy of predictions could be enhanced by either increasing the temporal resolution of the data, or including more meaningful connections between words in the predictive model. 

Subsequent work not only supported the hypothesis that individual words in a child's vocabulary are informative in predicting future vocabulary growth, but that predictions based on this knowledge reliably outperformed those based purely on demographics \cite{beckage_quantifying_2020}. It was also noted that the specific composition of a child's vocabulary significantly affects a model's ability to predict the child's future language acquisition (i.e. the acquisition of new words), suggesting that an individual's vocabulary has relevant and predictive information on the type of learner - and the learning trajectory - of a particular child.

\subsection{Overview of Graph Neural Network approaches}

Gori \cite{gori_new_2005} and Scarselli \cite{scarselli_graph_2005} introduced  \emph{Graph Neural Networks (GNN)} \cite{wu_surveyGNN_2021}, which extend the concept of conventional feedforward neural networks to graph-structured data. This development allowed for more effective processing of this type of data, which historically had posed challenges for many machine learning approaches.  Unlike traditional neural networks, which are often designed for handling data in regular structures, such as images or data sequences, GNNs process inherently irregular data with irregular structures, e.g. potentially varying graph sizes, with no fixed number of neighbours for each nodes, and with complex connectivity patterns and attributes. They manage relationships (edges) between entities (nodes/vertices) within a graph, by learning a vector representation of each node (i.e. a \emph{node embedding}) that not only depends on its own features but also on the features of its neighbours. For a vocabulary model, each node in the graph corresponds to a word, and is represented by a feature vector that is initialised with an attribute representing the probability that this word is known.

\begin{figure*}[t]
        \centering
        \includegraphics[width=0.8\textwidth]{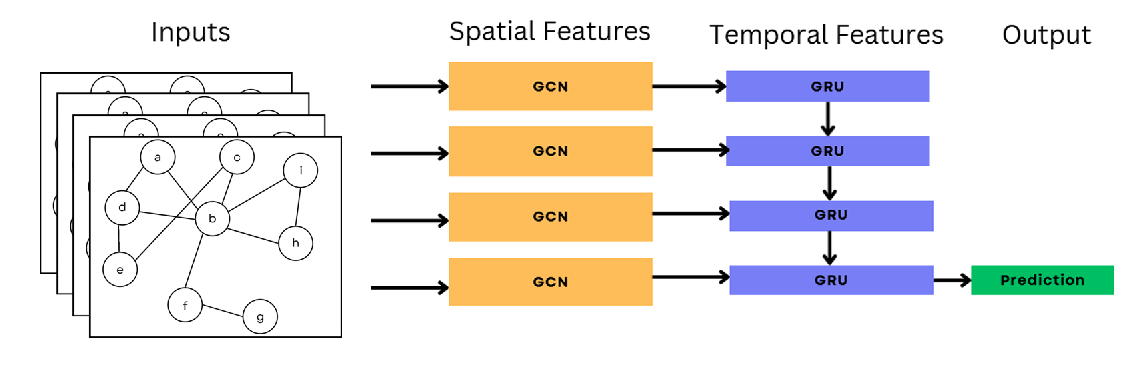}
        \caption{Block diagram of the T-GCN algorithm \cite{Zhao2020TGCN} used for the Spatio-Temporal Graph Convolutional Network model, using \emph{Graph Convolutional Networks} (GCNs) as spatial features and \emph{Gated recurrent units} (GRUs) as temporal features.}
        \label{fig:img2}
\end{figure*}

\emph{Spatial GNNs} focus on the spatial relationships between nodes, operating directly in the graph's node and edge space. They essentially `pass messages' between neighbouring nodes, aggregating information from a node's local neighbourhood to learn representations.  Fig.\ref{fig:img1} illustrates the general operation of a Spatial GNN \cite{roxburgh2024}; each node is selected and its local neighbourhood is determined (i.e. the \emph{Locality Phase}), which for a vocabulary model, is governed by the connection between word pairs. The graph structure is then reconfigured (i.e. the \emph{Aggregation Phase}) prior to transforming the aggregated features through a neural network layer resulting in a new feature vector for each node, that is informed by its neighbours (i.e. the \emph{Forward Propagation Phase}). Once all of the nodes have been processed, a final representation (i.e. a node embedding) is obtained for each of the nodes. These embeddings can then be used  to determine the probability that the words they represent are `known'.

\emph{Graph Convolutional Networks} (GCNs) \cite{Li_GCNsurvey_2022} apply a convolution operation to graphs in GNNs in a similar way to traditional Convolutional Neural Networks (CNNs), with many filtering methods possible, typically within the Fourier domain (Spectral GCNs) or in the spatial domain (Spatial GCNs). An interesting variation was proposed by Kipf \& Welling \cite{kipf_semi-supervised_2016} that merges the two methods, with spectral graph convolutions which are simplified to reduce the overhead that comes with computing a Fourier transform of a graph.

\emph{Spatio-Temporal~Graph~Convolutional~Networks} (STGCN) further extend GCNs, by combining a time dimension in addition to the spatial dimension, and as such, can be used for handling graph-structured, time-series data. Yan {\it et al.} \cite{yan2018spatial} first proposed the concept for recognising actions in skeleton models, and similar designs have been described as a method of predicting traffic in road networks \cite{yu_spatio_2018,Zhao2020TGCN}, with the data coming from road networks and traffic sensors represented as a graph, where edges represents direct routes between sensors, and the time dimension referring to the history of the sensor data \cite{jiang2022graph}. 

STGCN algorithms, such as T-GCN \cite{Zhao2020TGCN} shown in Fig. \ref{fig:img2},  typically employ a hybrid design that combines a graph convolutional network block, with a Recurrent Neural Network block such as a \emph{gated recurrent unit} (GRU), designed for sequential data. The graph convolutional network is used to capture the relationships between nodes, thus modelling spatial dependence, and the gated recurrent unit is used to capture the dynamics of the node features, modelling temporal dependence. STGCNs allow the modelling of evolving graph structures in which node dependencies change over time, which makes them a promising tool for attempting to predict a child's future vocabulary on the basis of a dynamic vocabulary together with fixed relationships between words.

%% %%%%%%%%%%%%%%%%%%%%%%%%%%%%%%%%%%%%%%%%%%%%%%%%%%%
\section{Modelling Vocabulary as a Graph}
\label{sec:model}
In a basic representation of the child's vocabulary as a graph, each node represents a word with an associated feature vector, incorporating information about the state of the word. We can represent the depth of the child's knowledge of each word by including a feature quantifying it. Each feature is initialised with discrete values that indicate that a child may:
\begin{itemize}
    \item 
i) understand a word but cannot produce it; 
    \item 
ii) produce a word but does not understand it; 
    \item 
iii) understand and produce a word; or 
    \item 
iv) have no knowledge of a word.
\end{itemize}
Whilst this model may be a possible oversimplification of a child's word knowledge and cognition, these discrete states have the advantage of being easily observable and well understood.

\begin{figure*}[t]
	\centering
	 
 	\includegraphics[width=0.65\linewidth]{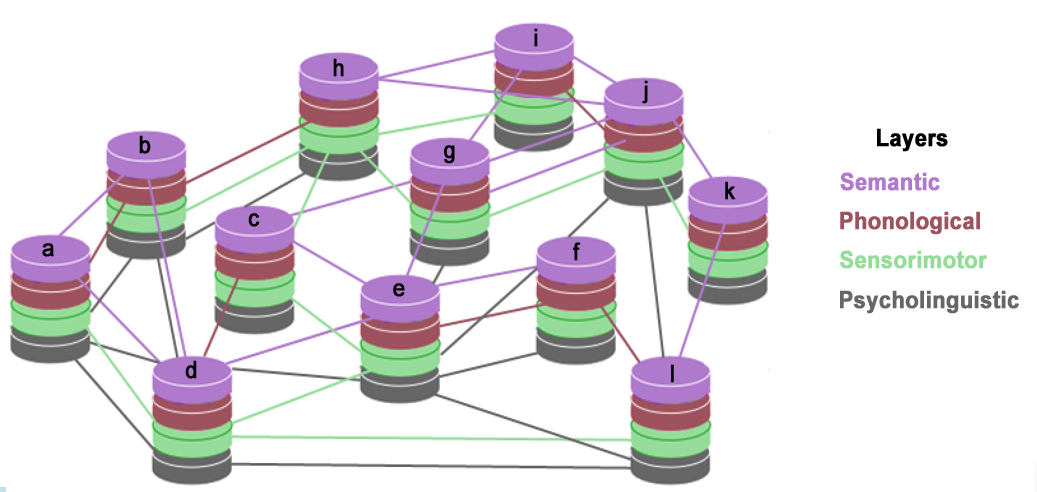}
    \caption {Simplified example of a \emph{Multiplex Lexical Network} with 4 relationship layers and 12 words represented by letters {\tt a - \tt l}. Edges are weighted (but not shown here).}
	\label{figure:multilayers}
\end{figure*}

The model can be enhanced by incorporating quantified relationships between words (thereby defining and weighting the graph edges). The various graphs can be arranged as a \emph{multiplex} \cite{wasserman_faust_1994} graph (Fig. \ref{figure:multilayers}), which is a type of graph where nodes represent the same set of items on all layers.  In the context of language, this has also been described as a \emph{Multiplex Lexical Network (MLN)} \cite{stella_multiplex_2017}. A variety of data exists that could be used for this purpose, including:
\begin{description}
    \item[Word Associations:]
        words that have some cognitive association with each other - for example \texttt{lock}/\texttt{key}, or \texttt{umbrella}/\texttt{rain};
    \item[Phonological Connections:]
        data on words that that have a measurable similarity due to sharing similar phonemes;
    \item[Psycholinguistic Norms:]
        how easy a word can be mentally visualised (i.e. \emph{Imageability}); 
    \item[Semantic Features:] 
        connecting words of a similar meaning in context; 
    \item[Concretedness:]
        whether a word represents a concrete or an abstract concept;
    \item[Familiariy:]
        how commonly used a word is; or 
    \item[Other:]
        simple meausres such as how difficult a word is to remember or pronounce (e.g. word length). 
\end{description}
Whilst a number of relationships exist that could be included, care needs to be taken not to rely too heavily on an `adult perspective', but rather concentrate on those more in line with an infant's cognitive perspective. In this paper, we discuss the incorporation of the following set of norms in the model:
\begin{description}
    \item[Semantic Feature Production norms] 
         measure the semantic similarity of different words.  We focus on using the McRae \cite{mcrae_semantic_2005} and Buchanan \cite{buchanan_english_2019} norms.
    \item[Sensorimotor norms] 
         measure perception and action strength of words from the perspective of the child.  The \emph{Lancaster Sensorimotor Norms} \cite{lynott2020lancaster} are used that evaluate English words based on six perceptual modalities (\emph{touch}, \emph{taste}, \emph{smell}, \emph{hearing},   \emph{vision}, and \emph{interoception}) and five action effectors (\emph{mouth/throat}, \emph{hand/arm}, \emph{foot/leg}, \emph{head excluding mouth/throat}, and \emph{torso}).
\end{description}

\begin{figure*}[t]
\centering
\begin{minipage}[c][][c]{.475\textwidth}
    \centering
    \includegraphics[width=\linewidth]{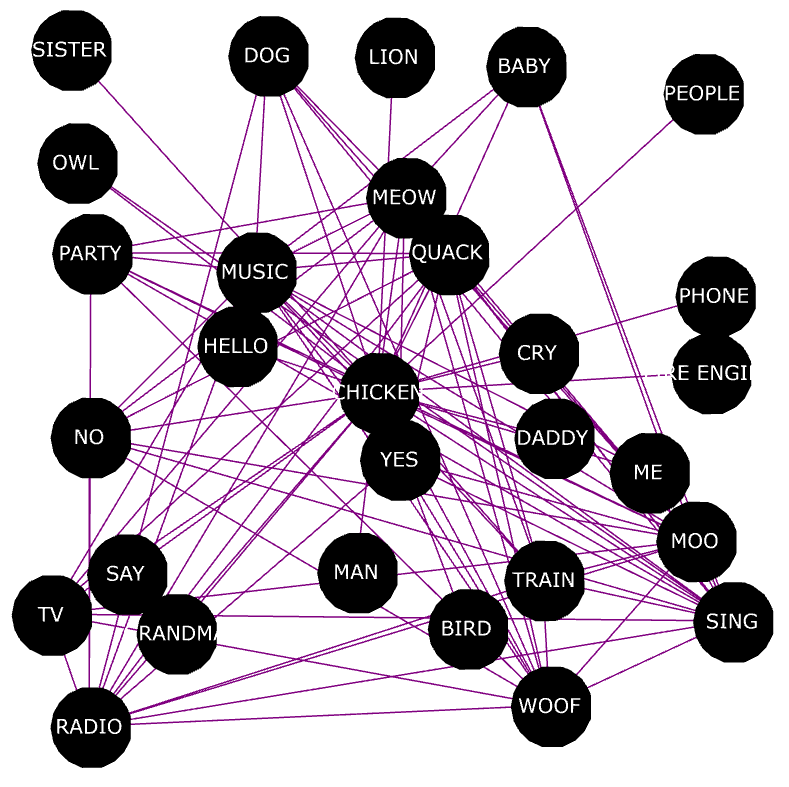}
    \caption{Example \emph{auditory sensorimotor} relationship graph (simplified) focused on the word \texttt{chicken} (based on Lancaster Sensorimotor Norms \cite{lynott2020lancaster}).}
    \label{fig:image1}
\end{minipage}~~~~~~~%
\begin{minipage}{.466\textwidth}
    \centering
    \includegraphics[width=\linewidth]{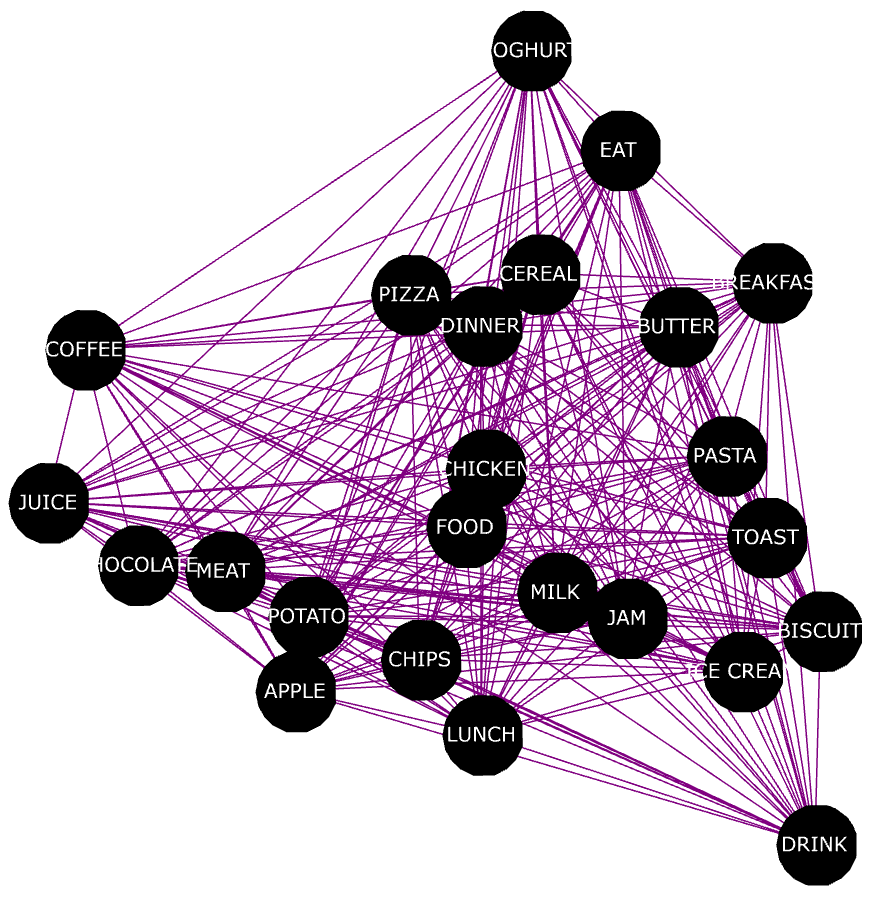}
    \caption{Example \emph{gustatory sensorimotor} relationship graph (simplified) focused on the word \texttt{chicken} (based on Lancaster Sensorimotor Norms \cite{lynott2020lancaster}).}
    \label{fig:image2}
\end{minipage}
        \vspace{-0.2in}
\end{figure*}

Sensorimotor norms are named after the initial phase in Piaget's theory of cognitive development \cite{piaget_origins_1953}, which spans from birth to approximately 2 years of age. During this period, infants learn to use their senses to construct an understanding of the world, and use motor movements (\emph{reaching}, \emph{sucking}, \emph{grasping}, and \emph{touching}) to interact with it.  Thus, the primary objective of the sensorimotor stage is to develop an understanding of \emph{object permanence}, i.e. the realisation that objects and events persist even when they are not directly observable by the child. From the perspective of infant language acquisition, it is especially useful to be able to connect words from the conceptual point of view of a child at the earliest development stage. Figs. \ref{fig:image1} and \ref{fig:image2} illustrate a simplified vocabulary graph with edges connecting nodes with similar \emph{Lancaster Norm} scores \cite{lynott2020lancaster}: Fig. \ref{fig:image1} illustrates nodes in an \emph{auditory} graph; whereas Fig. \ref{fig:image2} illustrates those in a \emph{gustatory} one. Both graphs are simplified and focused on the word \texttt{chicken}.

To represent the strengths of connections between words in the graph for each of the similarity measures we used, in the case of the Semantic Feature Production Norms, the necessary cosine similarity matrices were used together with the data by McRae and Buchanan. Lancaster norms require additional processing for use in our model, and thus a weighted adjacency matrix was created for all possible word pairs within each category by computing the cosine similarity of each pair, which in this case is simply the normalised product of their scores.  Thus, words with, for example, high scores in the \emph{Olfactory} category (e.g. \texttt{orange} and \texttt{lemon}) would have a strong connection, whilst a word with high score and a word with low score (e.g. \texttt{orange} and \texttt{table}) would have a weak connection. Similarly, the connection between two words with low scores in the \emph{Olfactory} category (e.g. \texttt{table} and \texttt{chair}) would also be weak. In addition, all self-loops (e.g. \texttt{head} and \texttt{head}) have a maximal weight (i.e.  1.0).

For each category, the resulting matrix comprised approximately 76,000 data points. Given that many of the pairs may not be relevant to the category and would result in an unnecessary computational burden, a threshold was used (with the inter-word connection value of 0.5) to eliminate pairs deemed unnecessary due to the implied lack of connection. For each sensorimotor category, we used the adjacency matrices to define the edges, and  created a list of nodes by removing duplicates in the edge list.  Finally, a feature was also added to each node representing the level of knowledge of that word.  

%% %%%%%%%%%%%%%%%%%%%%%%%%%%%%%%%%%%%%%%%%%%%%%%%%%%%
\section{Data Collection}
\label{sec:data}

Two types of data were collected to build a model and evaluate it; datasets of words and their relationships were collected in order to create the model itself (i.e. the \emph{Relational Datasets}), and observational data on child vocabulary was collected in order to test the resulting model (i.e. the \emph{Observational Dataset}). For the \emph{Relational Datasets}, we used the English language relationship datasets comprising both semantic and sensorimotor norms from \cite{buchanan_english_2019,lynott2020lancaster,mcrae_semantic_2005}. A challenge in putting together independent datasets into one model is one of standardisation.  This should be addressed by accounting for differences due to:
\begin{itemize}
    \item
i) synonyms (e.g. \texttt{rabbit} and \texttt{bunny});
    \item
ii) regional variations (e.g. \texttt{mommy} and \texttt{mummy}); and
    \item
iii) international spelling variations (e.g. \texttt{colour} and \texttt{color}). 
\end{itemize}
This was addressed by pre-processing the data to identify such cases, and renaming words for consistency, thus resulting in a \emph{standardised} vocabulary.

A second challenge was to consider \emph{homographs}, i.e. words with multiple meanings such as \texttt{back}, \texttt{bank}, \texttt{run}, or \texttt{drink}.  This was address by using a contextual specifier as a label for ambiguous words, thus ensuring that the vocabulary was unambiguous. For example, ambiguous words such as \texttt{drink} become \texttt{drink(beverage)} and \texttt{drink(verb)}. This is particularly challenging when considering the child perspective, as for certain words, such relationships are different than for adults.  To illustrate this, the notion of \texttt{fish} being a food and \texttt{fish} being an animal are typically treated as distinct concepts for children, whereas for adults \texttt{fish} is understood as simultaneously both a food and an animal. Again these words were appended with a context-appropriate label.

The \emph{Observational Dataset} consisted of data collected from participants using our web app as part of a larger study\footnote{The study has received ethical approval by the Faculty of Science and Engineering Research Ethics Committee, University of Liverpool, Ref. 8182.} as well as item-level CDI Survey responses extracted from all available forms in English acquired from Wordbank \cite{frank_wordbank_2017}. We selected only data for which there are longitudinal sequences, to enable the T-GCN algorithm's GRU layer block to train on temporal data. Words were converted to our standardised vocabulary to allow for Irish, Australian, American and regional British dialect differences. The overall dataset included 460 observations (i.e. vocabulary inventories), in sequences of 4 contiguous observations, using a sliding window method to create smaller sequences from longer ones. The training data consisted of 115 `test subjects' corresponding to individual sequences (as a child may generate more than one sequence), and the test dataset consisted of 120 observations across 30 `test subjects'.

Due to the nature of real-world, human-collected data regarding human behaviour, inevitably there will be errors present, and thus working with such data presents challenges. With a large dataset containing millions of observations, occasional errors will have minimal impact on the overall behaviour.  However, such errors may have an adverse effect on trials with smaller datasets (such as that used for this evaluation), and thus, to address this, the data was pre-processed to identify possible errors and eliminate discrepancies. Common errors encountered in observational data include:
\begin{description}
    \item[Observer Misinterpretation:] a child may be observed at one time period as understanding but not producing a particular word; however it becomes apparent over time that the child has a different interpretation for that word.
    \item[Mistaken Observation:] the parent believes that a child has said and/or understood a word, but they change their mind in the subsequent observation.
    \item[Observational Error:] a sequential pair of observations that made sense to the parent at the time of each observation, but that were subsequently confused, suggesting the child has either forgotten the meaning of the word, or forgotten how to say it.    In this study, we assume that children do not forget words once they have demonstrated knowledge of them.    
    \item[Unobserved knowledge:] the parent may be unaware that a child knows a particular word; e.g. the child may have used the word at a relatives' house.
    \item[Articulation Error:] the child may speak a certain word, but pronouncing it in such a way that it is unintelligible or ambiguous to an observer.
\end{description}

Dropping words from the observed vocabulary could adversely affect the ability to train a classifier on the data; therefore to deal with contradictory data, two datasets were generated, an {\em optimistic} one whereby it was assumed that the child continue to understand the word in subsequent observation periods, and a {\em pessimistic} one whereby the assumption is that words are false observations, and that the child did not understand the word during the first observation.

%% %%%%%%%%%%%%%%%%%%%%%%%%%%%%%%%%%%%%%%%%%%%%%%%%%%%
\section{Implementation and Evaluation}
\label{sec:results}

\begin{table*}[t]
    \def\arraystretch{1.1}
\centering
\caption{Experimental HyperParameters used by all STGCN models.}
\begin{tabular}{|l|c||l|c|}
\hline
\textbf{Parameter} & \textbf{Value} & \textbf{Parameter} & \textbf{Value} \\
\hline
\hline
Data Mode & ~~\emph{Optimistic}~~ & Epochs & \emph{1000} \\
\hline
Batch Size & \emph{4} & LSTM Sequence Length & \emph{4} \\
\hline
Prediction Length~~~~ & \emph{1} & Loss Optimiser & ~~\emph{ADAM}~~ \\
\hline
Loss Metric & \emph{MAE} & Model Metric & \emph{MSE} \\
\hline
Dropout & \emph{None} & Input Graph Connections Limit~~~~ & \emph{2000} \\
\hline
\end{tabular}
\label{tab:stgcn-hyperparameters}
\end{table*}

A collection of thirteen ($N=13$) individual models were used for the
vocabulary relationships explored in this study,
represented as graphs \(G_N = (V_N,E_N)\)  that were based on each of the norms described in the data model (Section \ref{sec:model}), where $V_i$ is the set of vertices or nodes, and $E_i$ are the corresponding set of edges for each model $i \in N$.
A set of nodes was created for every observation in the data, and was populated with the corresponding observed data.  These sets of nodes were combined to form a time series.
The set of edges were then processed by combining each edge set with each node set in the time series, resulting in the creation of a time series of graphs for each of the vocabulary relationships. The aim of the model is to predict a child's future vocabulary based on past and existing knowledge, represented as a series of graphs representing different relationships between words in a vocabulary over a number of discrete time periods.  Therefore, the nodes were embedded with feature vectors representing the child's current knowledge of the word at each time period. This comprehension attribute was assigned a starting value from one of four levels, reflecting the child's knowledge of that word at the given observation:
\begin{itemize}
    \item
no comprehension (0.0);
    \item 
production without understanding (0.3);
    \item 
understanding but no production (0.6); and
    \item 
full comprehension and production (1.0).
\end{itemize}

A Spatio-temporal Graph Convolutional Network-based model was developed in Python using the \emph{Stellargraph}\footnote{\url{https://github.com/stellargraph/stellargraph}} software library, and configured using the hyperparameters in Table \ref{tab:stgcn-hyperparameters}. Stellargraph is built on \emph{Tensorflow},\footnote{\url{https://www.tensorflow.org/}} and facilitates the construction of graph-based machine learning models. The Graph Neural Network algorithm used in this study is based on the Temporal-GCN (T-GCN) as described by Zhao et al. \cite{Zhao2020TGCN}. The full model consists of 13 relationship layers, each of which is a separate STGCN model that has been individually trained and executed.  Some nodes in these layers may be unconnected as they have no meaningful associations with other words in the context of that relationship.

When a new prediction is required, the vector representing the child's current vocabulary is used to populate the feature vectors of each node on each relationship layer, representing the new words that the child had recently learned.  The GCN classifier, in conjunction with the STGCN's spatio-temporal block, is then applied to these graphs to re-classify the unknown nodes.  These newly classified nodes can then be used to determine the words that are likely to be influenced the most by its neighbours, and so may be learned next.  This results in a list of candidate words from each relationship layer, from which the most likely words a child should learn can be determined.

\subsection{Training \& Validation}

The training stage of our STGCN involves presenting the model with a time series of graphs - analogous to a spatio-temporal graph - representing observations of childrens' vocabulary changing over time. This consists of a chain of 4-consecutive-observation chunks. The Sequence Length parameter of the STGCN was set to match the consecutive observation chunk size.  We split the input data into training, testing and validation sets.  

A \emph{2-layer Feedforward Neural Network} model was also trained using vocabulary data with no relationship element, to provide a baseline for the evaluation.
Each word in the child's vocabulary was linked to an input node in the network, which connects to a hidden layer with 500 units. Each word is represented as a node in the output layer.  
Other model hyperparameters include a Learning Rate of 0.8 and Momentum of 0.9, and the network was trained over 1000 epochs.

\begin{table*}[t]
    \def\arraystretch{1.1}
	\centering
    \caption{Results for each of the 13 individual models and the ANN baseline.}
	\begin{tabular}{|l|c|c|c|c|}
\hline
\textbf{Reference (baseline) Model} & \textbf{~Precision~} & \textbf{~Recall~} & \textbf{~~F1~~} & \textbf{~Accuracy~} \\ \hline 
~~~~~~~~~~2-Layer Feedforward (ANN) & \multirow{2}{*}{0.283} & \multirow{2}{*}{0.854} & \multirow{2}{*}{0.426} & \multirow{2}{*}{0.610} \\

~~~~~~~~~~Neural Network & & & & \\
\hline
\hline
\textbf{Semantic Relationships:} & & & &\\ \hline
~~~~~~~~~~McRae \cite{mcrae_semantic_2005}  & 0.450 & 0.513 & 0.479 & 0.740 \\ \hline
~~~~~~~~~~Buchanan \cite{buchanan_english_2019} & 0.403 & 0.606 & 0.484 & 0.715 \\
\hline 
\hline 
\textbf{Sensorimotor Relationships~\cite{lynott2020lancaster}:~~} & & & &\\ \hline

~~~~~~~~~~Lancaster (Haptic - \emph{Touch}) & 0.419 & 0.586 & 0.488 & 0.730 \\ \hline 
~~~~~~~~~~Lancaster (Gustatory - \emph{Taste}) & 0.427 & 0.598 & 0.498 & 0.731 \\ \hline
~~~~~~~~~~Lancaster (Olfactory - \emph{Smell}) & 0.424 & 0.571 & 0.487 & 0.733 \\ \hline
~~~~~~~~~~Lancaster (Auditory - \emph{Hearing}) & 0.465 & 0.395 & 0.427 & 0.750 \\ \hline
~~~~~~~~~~Lancaster (Visual - \emph{Vision}) & 0.438 & 0.618 & 0.513 & 0.732 \\ \hline
~~~~~~~~~~Lancaster (Interoceptive) & 0.435 & 0.494 & 0.462 & 0.739 \\ \hline
~~~~~~~~~~Lancaster (Mouth/Throat) & 0.417 & 0.637 & 0.504 & 0.716 \\ \hline
~~~~~~~~~~Lancaster (Hand/Arm) & 0.404 & 0.540 & 0.462 & 0.722 \\ \hline 
~~~~~~~~~~Lancaster (Foot/Leg) & 0.428 & 0.548 & 0.480 & 0.733 \\ \hline
~~~~~~~~~~Lancaster (Head) & 0.443 & 0.509 & 0.474 & 0.737 \\ \hline
~~~~~~~~~~Lancaster (Torso) & 0.438 & 0.537 & 0.483 & 0.741 \\ \hline

\end{tabular}
\label{tab:results}
\end{table*}

\subsection{Evaluation}
A total of thirteen models were evaluated (2 that model semantic relationships and 11 that model sensorimotor relationships, based on the norms described in Section \ref{sec:model}),
and compared with the baseline 2-layer Feedforward Neural Network. The aim is to evaluate each of these models independently, and compare them with the simple neural network approach.  The results are presented in Table \ref{tab:results}, together with the following evaluation metrics:
\begin{description}
    \item[Precision:] Measures the accuracy of positive predictions; i.e. out of all the positive predictions (i.e. words that the model has predicted will have increased comprehension), how many of those were proven to be correct.
        \[
        \text{Precision} = \frac{\text{True Positives}}{\text{True Positives} + \text{False Positives}}
        \]
    \item[Recall:] Measures the fraction of relevant instances retrieved; i.e. out of all the words that were actually proven to be learned, how many were correctly predicted by the model.
        \[
        \text{Recall} = \frac{\text{True Positives}}{\text{True Positives} + \text{False Negatives}}
        \]
    \item[F1 Score (or F1 Measure):] The harmonic mean of precision and recall. 
        \[
        F1 = 2 \times \frac{\text{Precision} \times \text{Recall}}{\text{Precision} + \text{Recall}}
        \]
    \item[Accuracy:] Measures the fraction of correct predictions, representing  the ratio of correctly predicted words with increased comprehension, to the total number of new words actually learned by the child.  
            \[
        \text{Accuracy} = \frac{\text{True Positives} + \text{True Negatives}}{\text{Total Predictions}}
        \]
\end{description}

Although the overall performance of the ANN demonstrated moderate performance, it fell short of that demonstrated by each of the STGCN models (Table \ref{tab:results}).  The results suggests that although many of the words were predicted, the percentage of those being correct was low. In comparison, the use of the different relationships with the STGCN model resulted in a greater accuracy of the words predicted, but with a drop in the overall percentage of possible predictions.

The accuracy of individual predictive models varied greatly despite achieving similar overall accuracy across all models for the entire test set. 
There is some variance in the performance of the different STGCN models incorporating Sensorimotor relationships, with \emph{Visual} and \emph{Mouth} models showing notably higher recall values at 0.618 and 0.637 respectively. 
These high recall values are indicative of the models' ability to identify a larger proportion of relevant words for learning, making them suitable choices for minimising false negatives.  
The \emph{Auditory} STGCN model clearly stands out with the highest accuracy (0.750) and highest F1-score (0.465) but the lowest Recall (0.395), suggesting it is more selective but might miss out on some words the child could learn.  
The \emph{Haptic}, \emph{Gustatory}, and \emph{Olfactory} models balance between F1-scores and Recall, with F1-scores in the 0.488 to 0.498 range, indicating a reasonable compromise between the metrics.  Overall, F1-scores for the Sensorimotor STGCN models are relatively high, ranging from 0.716 to 0.750, suggesting they make correct predictions about whether a word will or will not be typically learned. 
The \emph{Torso}, \emph{Foot/Leg}, and \emph{Interoceptive} models show neither highest nor lowest performance, but could still potentially provide useful predictions for vocabulary acquisition. 
The performance when using the semantic relationships was comparable to those of the sensorimotor relationships, with the McRae et al. norms resulting a model that performed marginally better than that using the Buchannan et al. norms.

A comparative evaluation was also conducted to compare the two variants of the observational data: \emph{optimistic} and \emph{pessimistic}, with the former correcting contradictions in observations by assuming the child knew the words that appeared to be forgotten, while the latter assumed an observational error by the caregiver and that the child did not know the word. In general, the \emph{optimistic} version outperformed the \emph{pessimistic} version, thus indicating the significance of the observational data in determining the accuracy of predictive models.

%% %%%%%%%%%%%%%%%%%%%%%%%%%%%%%%%%%%%%%%%%%%%%%%%%%%%
\section{Conclusions \& Discussion}
\label{sec:conclusions}
A Spatio-Temporal Graph Neural Network model (and the way in which it can be constructed) was presented that can be used to make predictions about a child's upcoming vocabulary acquisition. The model is based on existing work on infant language acquisition prediction using neural networks \cite{sims2013exploring} and graph models \cite{ke2008analysing}; however, we have expanded this by considering the current and past vocabularies of a given child combined with multiple types of relationships between the words, including sensorimotor norms \cite{lynott2020lancaster} and semantic feature prediction norms \cite{buchanan_english_2019,mcrae_semantic_2005}.

The overall accuracies of each model were reasonably similar; however on closer inspection,  individual prediction level showed several differences, suggesting there is scope for exploitation of the variation of coverage. One approach would be to combine and improve the predictability of multiple models through the use of an ensemble output stage \cite{hansen1990neural} that would allow each of the models to contribute towards the overall prediction.  
We will expand the number of relationships we use to inform the input graphs, including word association norms, phonological relationships, and psycholinguistic vocabulary norms such as imageability and concreteness.
Furthermore, given that this study addresses an important public health issue with an innovative methodology, a more detailed comparative analysis with existing state-of-the-art techniques could provide clearer insights into how well the model performs, as well as issues of scalability.

%%%%%%%%%%%%%%%%%%%%%%%%%%%%%%%%%%%%%%%%%%%%%%%%%%%%%%%%%%%%%%%%%%%%%%%%

%%% Use this environment to include acknowledgements (optional).
%%% This will be omitted in doubleblind mode.

%\begin{ack}
%Do we have acknowledgements?
%\end{ack}

%%%%%%%%%%%%%%%%%%%%%%%%%%%%%%%%%%%%%%%%%%%%%%%%%%%%%%%%%%%%%%%%%%%%%%%%

%%% Use this command to include your bibliography file.

%\bibliography{master_list}

\begin{thebibliography}{44}
\providecommand{\natexlab}[1]{#1}
\providecommand{\url}[1]{\texttt{#1}}
\expandafter\ifx\csname urlstyle\endcsname\relax
  \providecommand{\doi}[1]{doi: #1}\else
  \providecommand{\doi}{doi: \begingroup \urlstyle{rm}\Url}\fi

\bibitem[Alcock et~al.(2016)Alcock, Rowland, Meints, Christopher, Just, Brelsford, and Summers]{alcock_construction_2016}
K.~J. Alcock, C.~F. Rowland, K.~Meints, A.~E. Christopher, J.~Just, V.~Brelsford, and J.~Summers.
\newblock Construction and standardisation of the {UK} communicative development inventory {(UK-CDI)}, words and gestures.
\newblock In \emph{International Conference on Infant Studies}, 2016.

\bibitem[Baron-Cohen et~al.(2009)Baron-Cohen, Scott, Allison, et~al.]{baron-cohen_prevalence_2009}
S.~Baron-Cohen, F.~J. Scott, C.~Allison, et~al.
\newblock Prevalence of autism-spectrum conditions: {UK} school-based population study.
\newblock \emph{Br J Psychiatry}, 194\penalty0 (6):\penalty0 500--509, June 2009.

\bibitem[Beckage et~al.(2015)Beckage, Mozer, and Colunga]{beckage_predicting_2015}
N.~Beckage, M.~Mozer, and E.~Colunga.
\newblock Predicting a child's trajectory of lexical acquisition.
\newblock In D.~C. Noelle et~al., editors, \emph{37th Annual Meeting of the Cognitive Science Society, CogSci 2015}, 2015.

\bibitem[Beckage et~al.(2020)Beckage, Mozer, and Colunga]{beckage_quantifying_2020}
N.~M. Beckage, M.~C. Mozer, and E.~Colunga.
\newblock Quantifying the role of vocabulary knowledge in predicting future word learning.
\newblock \emph{IEEE Trans. Cogn. Develop. Syst.}, 12\penalty0 (2):\penalty0 148--159, June 2020.

\bibitem[Bleses et~al.(2016)Bleses, Makransky, Dale, et~al.]{bleses_early_2016}
D.~Bleses, G.~Makransky, P.~S. Dale, et~al.
\newblock Early productive vocabulary predicts academic achievement 10 years later.
\newblock \emph{Appl Psycholinguist}, 37\penalty0 (6):\penalty0 1461--1476, 2016.

\bibitem[Buchanan et~al.(2019)Buchanan, Valentine, and Maxwell]{buchanan_english_2019}
E.~M. Buchanan, K.~D. Valentine, and N.~P. Maxwell.
\newblock English semantic feature production norms: An extended database of 4436 concepts.
\newblock \emph{Behav Res Methods}, 51\penalty0 (4):\penalty0 1849--1863, Aug. 2019.

\bibitem[Clegg et~al.(2005)Clegg, Hollis, Mawhood, et~al.]{clegg_developmental_2005}
J.~Clegg, C.~Hollis, L.~Mawhood, et~al.
\newblock Developmental language disorders - a follow-up in later adult life. cognitive, language and psychosocial outcomes.
\newblock \emph{J. Child Psychol. Psychiatry}, 46\penalty0 (2):\penalty0 128--149, Feb. 2005.

\bibitem[Conti-Ramsden et~al.(2001)Conti-Ramsden, Botting, Simkin, et~al.]{conti-ramsden_follow-up_2001}
N.~Conti-Ramsden, Z.~Botting, E.~Simkin, et~al.
\newblock Follow-up of children attending infant language units: Outcomes at 11 years of age.
\newblock \emph{Int J Lang Commun Disord}, 36\penalty0 (2):\penalty0 207--219, 2001.

\bibitem[Feinstein and Duckworth(2006)]{feinstein_development_2006}
L.~Feinstein and K.~Duckworth.
\newblock Development in the early years: its importance for school performance and adult outcomes.
\newblock Technical report, Centre for Research on the Wider Benefits of Learning, London, 2006.

\bibitem[Fenson et~al.(2007)Fenson, Marchman, Thal, Dale, Reznick, and Bates]{fenson_macarthur-bates_2007}
L.~Fenson, V.~Marchman, D.~Thal, P.~Dale, J.~Reznick, and E.~Bates.
\newblock \emph{MacArthur-Bates Communicative Development Inventories}.
\newblock Paul H. Brookes Publishing Company, Baltimore, MD, 2nd edition, 2007.

\bibitem[Frank et~al.(2017)Frank, Braginsky, Yurovsky, et~al.]{frank_wordbank_2017}
M.~C. Frank, M.~Braginsky, D.~Yurovsky, et~al.
\newblock Wordbank: an open repository for developmental vocabulary data.
\newblock \emph{Journal of Child Language}, 44\penalty0 (3):\penalty0 677--694, 2017.

\bibitem[Gori et~al.(2005)Gori, Monfardini, and Scarselli]{gori_new_2005}
M.~Gori, G.~Monfardini, and F.~Scarselli.
\newblock A new model for learning in graph domains.
\newblock In \emph{Proceedings. 2005 IEEE International Joint Conference on Neural Networks, 2005.}, volume~2, pages 729--734, Montreal, Que., Canada, 2005. IEEE.

\bibitem[Hansen and Salamon(1990)]{hansen1990neural}
L.~Hansen and P.~Salamon.
\newblock Neural network ensembles.
\newblock \emph{IEEE Transactions on Pattern Analysis and Machine Intelligence}, 12\penalty0 (10):\penalty0 993--1001, 1990.

\bibitem[Hulme and Snowling(2013)]{hulme_developmental_2013}
C.~Hulme and M.~J. Snowling.
\newblock \emph{Developmental disorders of language learning and cognition}.
\newblock John Wiley \& Sons, Chichester, 2013.

\bibitem[Jiang and Luo(2022)]{jiang2022graph}
W.~Jiang and J.~Luo.
\newblock Graph neural network for traffic forecasting: A survey.
\newblock \emph{Expert Systems with Applications}, 207:\penalty0 117921, 2022.

\bibitem[Johnson and Jusczyk(2001)]{johnson_word_2001}
E.~K. Johnson and P.~W. Jusczyk.
\newblock Word segmentation by 8-month-olds: When speech cues count more than statistics.
\newblock \emph{J Mem Lang}, 44\penalty0 (4):\penalty0 548--567, May 2001.

\bibitem[Ke and Yao(2008)]{ke2008analysing}
J.~Ke and Y.~Yao.
\newblock Analysing language development from a network approach\textasteriskcentered.
\newblock \emph{Journal of Quantitative Linguistics}, 15\penalty0 (1):\penalty0 70--99, Feb. 2008.

\bibitem[Kipf and Welling(2017)]{kipf_semi-supervised_2016}
T.~Kipf and M.~Welling.
\newblock Semi-supervised classification with graph convolutional networks.
\newblock In \emph{5th Int Conf. on Learning Representations, {ICLR} 2017 (Poster)}, 2017.

\bibitem[Law et~al.(2011)Law, McBean, and Rush]{law2011communication}
J.~Law, K.~McBean, and R.~Rush.
\newblock Communication skills in a population of primary school-aged children raised in an area of pronounced social disadvantage.
\newblock \emph{Int J Lang Commun Disord}, 46\penalty0 (6):\penalty0 657--64, 2011.

\bibitem[Li et~al.(2022)Li, Liu, Yang, Peng, and Zhou]{Li_GCNsurvey_2022}
Z.~Li, F.~Liu, W.~Yang, S.~Peng, and J.~Zhou.
\newblock A survey of convolutional neural networks: Analysis, applications, and prospects.
\newblock \emph{IEEE Transactions on Neural Networks and Learning Systems}, 33\penalty0 (12):\penalty0 6999--7019, 2022.

\bibitem[Lindsay et~al.(2002)Lindsay, Dockrell, Mackie, et~al.]{lindsay_educational_2002-1}
G.~Lindsay, J.~Dockrell, C.~Mackie, et~al.
\newblock Educational provision for children with specific speech and language difficulties in {England and Wales}.
\newblock Technical report, CEDAR, University of Warwick, 2002.

\bibitem[Locke et~al.(2002)Locke, Ginsborg, and Peers]{locke_development_2002}
A.~Locke, J.~Ginsborg, and I.~Peers.
\newblock Development and disadvantage: implications for the early years and beyond.
\newblock \emph{Int. Journal of Language and Communication Disorders}, 37\penalty0 (1):\penalty0 3--15, Jan. 2002.

\bibitem[Lynott et~al.(2020)Lynott, Connell, Brysbaert, et~al.]{lynott2020lancaster}
D.~Lynott, L.~Connell, M.~Brysbaert, et~al.
\newblock The {Lancaster Sensorimotor Norms}: multidimensional measures of perceptual and action strength for 40,000 {English} words.
\newblock \emph{Behav Res Methods}, 52\penalty0 (3):\penalty0 1271--1291, 2020.

\bibitem[McRae et~al.(2005)McRae, Cree, Seidenberg, et~al.]{mcrae_semantic_2005}
K.~McRae, G.~S. Cree, M.~S. Seidenberg, et~al.
\newblock Semantic feature production norms for a large set of living and nonliving things.
\newblock \emph{Behavior Research Methods}, 37\penalty0 (4):\penalty0 547--559, Nov. 2005.

\bibitem[Piaget(1953)]{piaget_origins_1953}
J.~Piaget.
\newblock \emph{The Origins Of Intelligence In Children}.
\newblock Routledge, London, 1953.

\bibitem[Roulstone et~al.(2011)Roulstone, Law, Rush, et~al.]{roulstone_investigating_2011}
S.~Roulstone, J.~Law, R.~Rush, et~al.
\newblock Investigating the role of language in children's early educational outcomes.
\newblock Technical Report DFE-RR134, Department of Education, UK, 2011.

\bibitem[Roxburgh(2024)]{roxburgh2024}
A.~Roxburgh.
\newblock \emph{Using Graph Neural Networks to Predict Toddler Vocabulary Acquisition}.
\newblock PhD thesis, University of Liverpool, Liverpool, UK, 2024.

\bibitem[Roxburgh et~al.(2017)Roxburgh, Grasso, and Payne]{roxburgh2017}
A.~Roxburgh, F.~Grasso, and T.~Payne.
\newblock Predicting word learning to boost child language acquisition.
\newblock 7th Int Conf. on Digital Health, DH '17, London, UK, 2017.

\bibitem[Scarselli et~al.(2005)Scarselli, {Sweah Liang Yong}, Gori, et~al.]{scarselli_graph_2005}
F.~Scarselli, {Sweah Liang Yong}, M.~Gori, et~al.
\newblock Graph neural networks for ranking web pages.
\newblock In \emph{The 2005 IEEE/WIC/ACM International Conference on Web Intelligence (WI'05)}, pages 666--672. IEEE, 2005.

\bibitem[Scerri et~al.(2011)Scerri, Morris, Buckingham, Newbury, Miller, Monaco, Bishop, and Paracchini]{scerri_dcdc2_2011}
T.~S. Scerri, A.~P. Morris, L.-L. Buckingham, D.~F. Newbury, L.~L. Miller, A.~P. Monaco, D.~V. Bishop, and S.~Paracchini.
\newblock {DCDC2, KIAA0319 and CMIP} are associated with reading-related traits.
\newblock \emph{Biol Psychiatry}, 70\penalty0 (3):\penalty0 237--245, 2011.

\bibitem[Sims et~al.(2013)Sims, Schilling, and Colunga]{sims2013exploring}
C.~Sims, S.~Schilling, and E.~Colunga.
\newblock Exploring the developmental feedback loop: word learning in neural networks and toddlers.
\newblock In \emph{Proc. Annual Meeting of the Cognitive Science Society, CogSci 2013}, volume~35, pages 3408--3413, 2013.

\bibitem[Snowling et~al.(2001)Snowling, Adams, Bishop, et~al.]{snowling_educational_2001}
M.~J. Snowling, J.~W. Adams, D.~V. Bishop, et~al.
\newblock Educational attainments of school leavers with a preschool history of speech-language impairments.
\newblock \emph{International Journal of Language and Communication Disorders}, 36\penalty0 (2):\penalty0 173--183, 2001.

\bibitem[Stadthagen-Gonzalez and Davis(2006)]{stadthagen2006bristol}
H.~Stadthagen-Gonzalez and C.~J. Davis.
\newblock The {Bristol} norms for age of acquisition, imageability, and familiarity.
\newblock \emph{Behav Res Methods}, 38\penalty0 (4):\penalty0 598--605, 2006.

\bibitem[Stella et~al.(2017)Stella, Beckage, and Brede]{stella_multiplex_2017}
M.~Stella, N.~M. Beckage, and M.~Brede.
\newblock Multiplex lexical networks reveal patterns in early word acquisition in children.
\newblock \emph{Scientific Reports}, 7\penalty0 (March):\penalty0 1--10, 2017.
\newblock ISSN 20452322.

\bibitem[Stothard et~al.(1998)Stothard, Snowling, Bishop, et~al.]{stothard_language_1998}
S.~E. Stothard, M.~J. Snowling, D.~V.~M. Bishop, et~al.
\newblock Language-impaired preschoolers.
\newblock \emph{Journal of Speech Language and Hearing Disorders}, 41\penalty0 (2):\penalty0 407--418, 1998.

\bibitem[Tomblin et~al.(1997)Tomblin, Records, Buckwalter, Zhang, Smith, and O'Brien]{tomblin_prevalence_1997}
J.~B. Tomblin, N.~L. Records, P.~Buckwalter, X.~Zhang, E.~Smith, and M.~O'Brien.
\newblock Prevalence of specific language impairment in kindergarten children.
\newblock \emph{Journal of Speech Language and Hearing Disorders}, 40\penalty0 (6):\penalty0 1245--1260, Dec. 1997.

\bibitem[Van~Dulm and Southwood(2016)]{van_dulm_does_2016}
O.~Van~Dulm and F.~Southwood.
\newblock Does socioeconomic level have an effect on school-age language skills in a developed country?
\newblock \emph{Stellenbosch Papers In Linguistics Plus}, 49, Dec. 2016.

\bibitem[Walker et~al.(1994)Walker, Greenwood, Hart, et~al.]{walker_prediction_1994}
D.~Walker, C.~Greenwood, B.~Hart, et~al.
\newblock Prediction of school outcomes based on early language production and socioeconomic factors.
\newblock \emph{Child Development}, 65\penalty0 (2):\penalty0 606, Apr. 1994.

\bibitem[Wasserman and Faust(1994)]{wasserman_faust_1994}
S.~Wasserman and K.~Faust.
\newblock \emph{Social Network Analysis: Methods and Applications}.
\newblock Structural Analysis in the Social Sciences. Cambridge University Press, 1994.

\bibitem[Wu et~al.(2021)Wu, Pan, Chen, Long, Zhang, and Yu]{wu_surveyGNN_2021}
Z.~Wu, S.~Pan, F.~Chen, G.~Long, C.~Zhang, and P.~S. Yu.
\newblock A comprehensive survey on {Graph Neural Networks}.
\newblock \emph{IEEE Transactions on Neural Networks and Learning Systems}, 32\penalty0 (1):\penalty0 4--24, 2021.

\bibitem[Yan et~al.(2018)Yan, Xiong, and Lin]{yan2018spatial}
S.~Yan, Y.~Xiong, and D.~Lin.
\newblock Spatial temporal graph convolutional networks for skeleton-based action recognition.
\newblock In \emph{Proceedings of the AAAI conference on Artificial Intelligence}, volume 32(1), pages 7444--7452, 2018.

\bibitem[Young et~al.(2002)Young, Beitchman, Johnson, et~al.]{young_young_2002}
A.~R. Young, J.~H. Beitchman, C.~Johnson, et~al.
\newblock Young adult academic outcomes in a longitudinal sample of early identified language impaired and control children.
\newblock \emph{Journal of Child Psychology and Psychiatry}, 43\penalty0 (5):\penalty0 635--645, July 2002.

\bibitem[Yu et~al.(2018)Yu, Yin, and Zhu]{yu_spatio_2018}
B.~Yu, H.~Yin, and Z.~Zhu.
\newblock Spatio-temporal graph convolutional networks: A deep learning framework for traffic forecasting.
\newblock In \emph{Proceedings of the Twenty-Seventh International Joint Conference on Artificial Intelligence (IJCAI)}, 2018.

\bibitem[Zhao et~al.(2020)Zhao, Song, Zhang, Liu, Wang, Lin, Deng, and Li]{Zhao2020TGCN}
L.~Zhao, Y.~Song, C.~Zhang, Y.~Liu, P.~Wang, T.~Lin, M.~Deng, and H.~Li.
\newblock {T-GCN}: A temporal graph convolutional network for traffic prediction.
\newblock \emph{IEEE Transactions on Intelligent Transportation Systems}, 21\penalty0 (9):\penalty0 3848--3858, 2020.

\end{thebibliography}

\end{document}